\documentclass[11pt,a4paper]{article}
\usepackage[breaklinks]{hyperref}
\usepackage[hyperref]{emnlp-ijcnlp-2019}

\usepackage[T1]{fontenc}
\usepackage[utf8]{inputenc}
\usepackage{times}
\usepackage{latexsym}
\usepackage{multirow}
\usepackage{multicol}
\usepackage[hyphenbreaks]{breakurl}
\usepackage{url}

\usepackage{todonotes}
\usepackage{textcomp}

\aclfinalcopy 

\title{Improving Neural Machine Translation Robustness via Data Augmentation: Beyond Back Translation}

\author{Zhenhao Li \\
  Department of Computing \\
  Imperial College London, UK \\
  {\tt zhenhao.li18@imperial.ac.uk} \\\And
  Lucia Specia \\
  Department of Computing \\
  Imperial College London, UK \\
  {\tt l.specia@imperial.ac.uk} \\}

\date{}

\begin{document}
\maketitle
\begin{abstract}
Neural Machine Translation (NMT) models have been proved strong when translating clean texts, but they are very sensitive to noise in the input. Improving NMT models robustness can be seen as a form of ``domain'' adaption to noise. The recently created Machine Translation on Noisy Text task corpus provides noisy-clean parallel data for a few language pairs, but this data is very limited in size and diversity. The state-of-the-art approaches are heavily dependent on large volumes of back-translated data. 
This paper has two main contributions: Firstly, we propose new data augmentation methods to extend limited noisy data and further improve NMT robustness to noise while keeping the models small. Secondly, we explore the effect of utilizing noise from external data in the form of speech transcripts and show that it could help robustness.
\end{abstract}

\section{Introduction}
Neural Machine Translation (NMT) models trained on large, clean parallel corpora have reached impressive performance in translating clean texts following various architectures \cite{DBLP:journals/corr/BahdanauCB14, gehring2017convolutional, vaswabu2017attention}. 

Despite this success, NMT models still lack robustness when applied to noisy sentences. \newcite{Belinkov2017SyntheticAN} show that perturbations in characters could cause a significant decrease in translation quality. They point out that training on noisy data, which can be seen as adversarial training, might help to improve model robustness. \newcite{michel-neubig-2018-mtnt} propose the Machine Translation on Noisy Text (MTNT) dataset, which contains parallel sentence pairs with comments crawled from Reddit\footnote{\url{www.reddit.com}} and manual translations. This dataset contains user-generated text with different kinds of noise, e.g., typos, grammatical errors, emojis, spoken languages, etc. for two language pairs.

In the WMT19 Robustness Task\footnote{\url{http://www.statmt.org/wmt19/robustness.html}} \cite{li-EtAl:2019:WMT1}, improving NMT robustness is treated as a domain adaption problem. The MTNT dataset is used as in-domain data, where models are trained with clean data and adapted to noisy data. Domain adaption is conducted in two main methods: fine tuning on in-domain data \cite{dabre-sumita:2019:WMT, post-duh:2019:WMT} and mixed training with domain tags \cite{berard-calapodescu-roux:2019:WMT, zheng-EtAl:2019:WMT}. The size of the noisy data provided by the shared task is small, with only thousands of noisy sentence pairs on each direction. Hence most approaches participating in the task performed noisy data augmentation using back translation \cite{berard-calapodescu-roux:2019:WMT, helcl-libovick-popel:2019:WMT, zheng-EtAl:2019:WMT}, with some approaches also directly adding synthetic noise \cite{berard-calapodescu-roux:2019:WMT}. The robustness of an NMT model can be seen as denoising source sentences (e.g. dealing with typos, etc.) while keeping a similar level of informal language in the translations (e.g. keeping emojis/emoticons). Based on this assumption, we believe that back translation of clean texts, although providing a large volume of  extra data, is limited since it removes most types of noise from the translations. In addition to adapting models on noisy parallel data, other techniques have been used to improve performance, generally measured according to BLEU \cite{papineni-etal-2002-bleu} against clean references. For example, \newcite{berard-calapodescu-roux:2019:WMT} apply inline-casing by adding special tokens before each word to represent word casing. In \cite{murakami-EtAl:2019:WMT}, placeholders are used to help to translate sentences with emojis. 

In this paper, we also explore data augmentation for robustness but focus on techniques other than back translation. We follow the WMT19 Robustness Task and conduct experiments under constrained and unconstrained data settings on Fr$\leftrightarrow$En as language pairs. Under the constrained setting, we only use datasets provided by the shared task, and propose new data augmentation methods to generate noise from this data. We compare back translation (BT) \cite{sennrich-etal-2016-improving} with forward translation (FT) on noisy texts and find that pseudo-parallel data from forward translation can help improve more robustness. We also adapt the idea of fuzzy matches from \newcite{bulte-tezcan-2019-neural} to the MTNT case by finding similar sentences in a parallel corpus to augment the limited noisy data. Results show that the fuzzy match method can extend noisy parallel data and improve model performance on both noisy and clean texts. The proposed techniques substantially outperform the baseline. While they still lag behind the winning submission in the WMT19 shared task, the resulting models are trained on much smaller clean data but augmented noisy data, leading to faster and more efficient training. Under the unconstrained setting, we propose for the first time the use of speech datasets, in two forms: (a) the IWSLT \cite{cettoloEtAl:EAMT2012} and MuST-C \cite{di-gangi-etal-2019-must} human transcripts as a source of spontaneous, clean speech data, and (b) automatically generated transcripts for the MuST-C dataset as another source of noise. We show that using informal language from spoken language datasets can also help to increase NMT robustness.

This paper is structured as follows: Section 2 introduces the data augmentation methods for noisy texts, including the previously proposed methods and our approaches to data augmentation. Section 3 describes our experimental settings, including the datasets we used, the augmented data and the baseline model. Section 4 shows the results of models built from  different training and evaluated on both clean and noisy test sets.   

\section{Noisy Data Augmentation}
\subsection{Previous Work}
Considering the limited size of noisy parallel data, data augmentation methods are commonly used to generate more noisy training materials.

Previous methods include back translation, injecting synthetic noise, and adversarial attacks. In the WMT19 Robustness Task, back translation on monolingual data was used to generate noisy parallel sentences \cite{murakami-EtAl:2019:WMT, zhou-EtAl:2019:WMT, berard-calapodescu-roux:2019:WMT, helcl-libovick-popel:2019:WMT}. \newcite{zheng-EtAl:2019:WMT} proposed an extension of back translation that generates noisy translations from clean monolingual data. Therefore, after reversing the direction, the noisy translations become the source, which would simulate the noisy source sentences from the MTNT parallel data. Synthetic noise is injected into clean data to form noisy parallel data in \newcite{Belinkov2017SyntheticAN, karpukhin2019training}. However, rule-based synthetic noise injection is limited to certain types of noise. Adversarial methods are proposed to inject random noise into clean training data in \newcite{cheng-etal-2018-towards, cheng-etal-2019-robust}.

We explore the following new methods as alternative ways to augment noisy parallel data.

\subsection{Fuzzy Matches} 
We adapted the method to augment data from parallel corpus from \newcite{bulte-tezcan-2019-neural}. The original method aims to find similar source sentences to those in a parallel corpus $(S, T)$ using a monolingual corpus, and then reuse the translation of the original source sentences as translations for the similar source sentences found. We adapted it to use only on the provided noisy training corpus. For each source sentence $s_i\in S$ in the training set, all other source sentences $s_j\in S(s_i\neq s_j)$ are compared with this sentence by measuring string similarity $Sim(s_i,s_j)$. If the similarity of the two sentences is above a threshold $\lambda$, the two sentences are mapped to each other's corresponding target sentence and the two new sentence pairs $(s_i, t_j), (s_j,t_i)$ are added into our augmented training data. The similarity is measured with Levenshtein distance \cite{levenshtein1966binary} on the token level. The similarity score is calculated as the edit distance divided by the minimum length of the two sentences (Equation \ref{eq:sim_score}). In addition to fuzzy matches in the parallel corpus, we  experimented with the monolingual corpus by mapping sentence $m_i$ in monolingual corpus to its fuzzy match's target sentence $t_j$ (If $Sim(m_i, s_j) > \lambda$, we add new sentence pair $(m_i, t_j)$ to the training augmented data). 
\begin{equation}
    Sim(S_i,S_j) = \frac{editdistance(s_i,s_j)}{min(len(s_i), len(s_j))}
    \label{eq:sim_score}
\end{equation}
To boost the speed of finding matches, we followed the  approach in \newcite{bulte-tezcan-2019-neural} and used a Python library \verb|SetSimilaritySearch|\footnote{\url{https://github.com/ekzhu/SetSimilaritySearch}} to select similar candidates before calculating edit distance. For each source sentence, only the top 10 similar candidates are selected to calculate the edit distance score. 

\subsection{Forward Translation}
Back translation  \cite{sennrich-etal-2016-improving} is a very popular technique for in-domain data augmentation. In the experiment, we back-translated MTNT monolingual data using a model fine-tuned on MTNT parallel corpus. However, considering that the task of improving robustness is to produce less noisy output, data generated with back translation might have noisy target translations (from monolingual data) and less noisy source texts (from back translation). Since this might increase the noise level of the output texts, we also experimented with forward translation using models fine-tuned on the noisy parallel corpus. Pseudo parallel data generated by forward translation is used for fine tuning models on the same language direction. To avoid overfitting, we merged the noisy parallel data of both language directions to produce noisy forward translations. The pseudo parallel data generated by back translation and forward translation is combined with noisy parallel data and fine-tuned on the baseline model. 

\subsection{Automatic Speech Recognition}
\label{ASR}
We used ASR systems and transcribed audio files into texts. In this case, we would expect noise to be generated during the process of automatic speech recognition. We selected a dataset with both audio and human transcripts, namely the MuST-C dataset. In this dataset, audio files $A$, human transcripts $S$ and transcript translated into another language $T$ are provided. We used Google Speech-to-Text API\footnote{\url{https://cloud.google.com/speech-to-text/}} and transcribed the audio files into automatic transcripts $S^\prime$. The human and ASR transcripts of the audio ($S$ and $S^\prime$) are treated as the source texts while the translations $T$ are target texts. We formed a new set of parallel data $(S^\prime,T)$ with ASR generated texts and the corresponding gold translations by humans.

Looking into the ASR transcripts, we found that the ASR system tends to skip some sentences due to the fast speaking speed, therefore we did some filtering based on the length ratio of the human transcripts and the ASR transcripts. We set a ratio threshold $\lambda$. For each pair $s_i, t_i$ in the ASR parallel data $(S^\prime, T)$, we removed this sentence pair if the length of $t_i$ dividing by $s_i$ is larger than the threshold $\lambda$.

\section{Experiments}
\label{sec:experiments}
\subsection{Corpora}
We used all parallel corpora from the WMT19 Robustness Task on Fr$\leftrightarrow$En. For out-of-domain training, we used the WMT15 Fr$\leftrightarrow$En News Translation Task data, including Europarl v7, Common Crawl, UN, News Commentary v10, and Gigaword Corpora. In the following sections, we represent the combination of these corpora as ``clean data". The MTNT dataset is used as our in-domain data for fine tuning. We also experimented with external corpora, namely the IWSLT2017\footnote{\url{https://wit3.fbk.eu/mt.php?release=2017-01-trnted}} and MuST-C\footnote{\url{https://ict.fbk.eu/must-c/}} corpora\footnote{The data from IWSLT has the same sentences in both translation directions, so we reversed the En$\rightarrow$Fr data on the Fr$\rightarrow$En direction. The MuST-C data is only used on En$\rightarrow$Fr direction.}, to explore the effect of informal spoken languages in human transcripts from speech. We also utilized monolingual data in the MTNT dataset. The size of parallel and monolingual data is shown in Table \ref{tab:corproa-size}. 

We used the development set in MTNT and the \textit{newsdiscussdev2015} for validation. Models with best performance on the validation set are evaluated on both noisy (MTNT and MTNT2019) and clean (\textit{newstest2014} and \textit{newsdiscusstest2015}) test sets. 

For prepossessing, we first tokenized the data with Moses tokenizer \cite{koehn-etal-2007-moses} and applied Byte Pair Encoding (BPE) \cite{sennrich-etal-2016-neural} to segment words into subwords. We experimented with a large vocabulary to include noisy as well as clean tokens and applied 50k merge operations for BPE. Upon evaluation, we detokenized our hypothesis files with the Moses detokenizer. We used \verb|multi-bleu-detok.perl| to evaluate the BLEU score on the test sets.

\begin{table}[htb]
\centering
\resizebox{\columnwidth}{!}{
\begin{tabular}{l c c c}
\multirow{2}{*}{\textbf{Corpus}} & \multirow{2}{*}{\textbf{Sentences}} & \multicolumn{2}{c}{\textbf{Words}} \\ 
& & \textbf{source} & \textbf{target} \\ \hline
Gigaword & 22.52M & 575.67M & 672.07M\\ \hline
UN Corpus & 12.89M & 316.22M & 353.92M \\ \hline
Common Crawl & 3.24M & 70.73M & 76.69M \\ \hline
Europarl & 2.01M & 50.26M & 52.35M \\ \hline
News Commentary & 200k & 4.46M & 5.19M \\ \hline
IWSLT       & 236k      & 4.16M     & 4.34M\\ \hline
MuST-C(en-fr)   & 275k      & 5.09M     & 5.30M\\ \hline
MTNT(en-fr) & 36k    & 841k      & 965k\\ \hline
MTNT(fr-en) & 19k    & 661k      & 634k\\ \hline
MTNT(en)    & 81k   & 3.41M  & \\ \hline
MTNT(fr)    & 26k   & 1.27M  & \\
\end{tabular}
}
\caption{Size of parallel and monolingual data. Source language represents English unless specified in brackets. The last two rows are noisy monolingual datasets in the MTNT dataset.}
\label{tab:corproa-size}
\end{table}

\subsection{Augmented Data}
\textbf{Fuzzy Match}
Considering the large size of our clean data, we only looked for noisy matches within the MTNT parallel and monolingual data. The parallel data on both directions was merged to get more combinations. With the similarity threshold setting to 50$\%$, we have 7,290 new sentence pairs in the En$\rightarrow$Fr and 7,154 in the Fr$\rightarrow$En language directions.
\\
\\
\textbf{Forward Translation}
We used noisy monolingual data to produce forward translations. Therefore, the number of parallel samples is the same as the monolingual data in Table \ref{tab:corproa-size}. Apart from tokenization and BPE, we did not do any other preprocessing to the forward translation.
\\
\\
\textbf{ASR}
We filtered the ASR parallel data by different length ratio thresholds (1.5 and 1.2). We measured the noise level of ASR transcripts $S^\prime$ by evaluating Word Error Rate (WER) and Word Recognition Rate (WRR) with \verb|asr_evaluation|\footnote{\url{https://github.com/belambert/asr-evaluation}} library, comparing it to the gold human transcripts $S$. 
\begin{table}[htb]
\centering
\begin{tabular}{|l|c|c|}
\hline
\textbf{Data} & \textbf{WER} & \textbf{WRR}\\ \hline
ASR($\lambda=1.5$) & 36.41\% & 65.38\% \\ \hline
ASR($\lambda=1.2$) & 31.70\% & 70.54\% \\ \hline
\end{tabular}
\caption{Noise level of ASR generated data filtered with different length ratio thresholds.}
\label{tab:asr-wer}
\end{table}

\begin{table*}[thb]
    \centering
    \begin{tabular}{ l l c c c c}
        \hline
        \textbf{Models} & \textbf{Fine-tune} & \textbf{MTNT} & \textbf{MTNT2019} & \textbf{newstest2014} &   \textbf{newsdiscusstest2015}\\ \hline
        Baseline   & None &  34.41   &   36.14   &	\textbf{36.51}	&   34.43   \\ \hline
        \multirow{5}{*}{Constrained}& Tune-S      & 40.16 &	41.58 &	35.94 &	\textbf{36.75}   \\ 
        & \quad+BT   & 37.93	&   39.19 &	34.03 &	34.00      \\ \cline{2-6}
        & Tune-B  & 38.25 &	40.12 &	35.36 &	35.15   \\
        & \quad+FM & 38.91 &	40.85 &	35.62 &	34.58 \\
        & \quad+FT & 39.38 &	41.82 &	35.56 &	35.00\\
        & \quad+FT+FM & 40.13 &	\textbf{42.80} &	35.82 &	35.88   \\
        & \quad\quad+double tune & \textbf{40.57} &	42.55 &	36.06 &	36.53 \\ \hline
        \multirow{2}{*}{Unconstrained} & IWSLT & 34.22    &	38.28 &	35.96 &	32.95 \\
         & MTNT+IWSLT & 37.52   & 41.22 & 36.33 &   34.07   \\ \hline     
    \end{tabular}
    \caption{BLEU scores of models fine-tuned on different data in the Fr$\rightarrow$En direction. The Tune-B model is fine-tuned with MTNT data merging both Fr$\rightarrow$En and En$\rightarrow$Fr directions while the Tune-S is fine-tuned only with Fr$\rightarrow$En data. BT and FT stand for back and forward translation, respectively. FM means fuzzy matching data. By double tuning we mean we fine-tune the model a second time on the MTNT training data.}
    \label{table:bleu_fren}
\end{table*}
\begin{table*}[thb]
    \centering
    \resizebox{\textwidth}{!}{
    \begin{tabular}{ l l c c c c}
        \hline
        \textbf{Models} & \textbf{Fine-tune} & \textbf{MTNT} & \textbf{MTNT2019} & \textbf{newstest2014} &   \textbf{newsdiscusstest2015}\\ \hline
        Baseline                    & None &  30.12 &	29.57 &	35.51 &	35.93   \\ \hline
        \multirow{4}{*}{Constrained}& MTNT(Tune-S)  & 36.15 &	32.25 &	36.77 &	37.16   \\ 
        & \quad+fix punctuation & --- & 35.49 & --- & --- \\ \cline{2-6}
        & MTNT$^*$(Tune-B) & 35.64 &	31.27 &	36.67 &	37.21 \\
        & \quad +FM & 36.03 & 31.37 & 36.58 & 37.13 \\
        & \quad +FT & 35.98 & 31.37 &	36.30 & \textbf{37.91} \\
        & \quad+FT+FM   & 36.21 &	31.25 &	36.37 &	37.76 \\
        & \quad\quad+double tune & \textbf{36.78} &	32.10 &	36.72 &	37.84 \\
        & \quad\quad\quad+fix punctuation & --- & \textbf{36.46} & --- & --- \\ \hline
        \multirow{4}{*}{Unconstrained} & IWSLT & 33.66 &	30.58 &	\textbf{36.89} &	37.34 \\
        & MTNT+IWSLT & 35.44 &	31.24 &	36.73 &	37.90 \\
        & ASR($\lambda=1.5$) &  30.53 &	28.66 &	36.46 &	35.61\\
        & ASR($\lambda=1.2$) & 31.09 &	29.48 &	36.59 &	35.54 \\
        & MuST-C & 34.15 &	31.09 &	36.27 &	37.61  \\\hline
        Mixed training & None & 35.32 &	31.14 &	35.5 &	36.05 \\ \hline
    \end{tabular}
    }
    \caption{BLEU scores of models fine-tuned on different data in the En$\rightarrow$Fr direction. $\lambda$ in ASR data represents the filtering threshold, as mentioned in Section \ref{ASR}. The mixed-training model combines all data available and adds domain tags in front of each sentence. Other notations are same as in Table \ref{table:bleu_fren}.}
    \label{table:bleu_enfr}
\end{table*}

\subsection{Model}
Our baseline model, which only uses clean data for training, is the standard Transformer model \cite{vaswabu2017attention} with default hyper-parameters. The batch size is 4096 tokens (subwords). Our models are trained with OpenNMT-py \cite{opennmt} on a single GTX 1080 Ti for 5 epochs. We experimented with fine tuning on noisy data and mixed training with ``domain'' tags \cite{caswell-chelba-grangier:2019:WMT, kobus2016domain} indicating where the sentences are sourced from. We used different tags for clean data, MTNT parallel data, forward translation, back translation, ASR data, and fuzzy match data. Tags are added at the beginning of each source sentences.

During the fine tuning on in-domain data, we continued the learning rate and model parameters. We stop the fine tuning when the perplexity on noisy validation set does not improve after 3 epochs. Best fine-tuned checkpoints are evaluated on the test sets.

The MTNT dataset provides noisy parallel data in specific language pairs. We used models with two fine tuning strategies:  Tune-S and Tune-B. The Tune-S model is fine-tuned only with the noisy parallel data in the same direction while the Tune-B model is fine-tuned with the combination of both language directions (Fr$\rightarrow$En and En$\rightarrow$Fr).

\section{Results}
We evaluated the models fine-tuned on different datasets in terms of BLEU  on both noisy and clean test sets. We note that although both MTNT and MTNT2019 test sets are noisy, the MTNT2019 is less noisy and contains fewer occurrences of noise such as emojis. Similarly, since the \textit{newsdiscuss} test set contains informal language, it is slightly noisier than \textit{newstest} test set. The evaluation results for both directions are shown in Tables \ref{table:bleu_fren} and \ref{table:bleu_enfr}. 

\subsection{Fine Tuning on Noisy Text}
It can be seen that for both directions fine tuning on noisy data gives better performance for the noisy test set. Although the size of training data in MTNT is only 19k and 36k sentences, by simply fine tuning on it, the BLEU scores of Tune-S model increase by +5.65 and +6.03 on MTNT test set, +5.44 and +2.68 on MTNT2019 test set (See the second row in the tables). It is also worth noticing that although fine-tuned on noisy data, the performance on clean test sets increases as well. This shows that noisy parallel data could improve model robustness on both noisy and clean texts.

\subsection{Data Augmentation}
As it is common in the field, we experimented with back translation (third row in Table \ref{table:bleu_fren}). We used the target-to-source Tune-S model to back-translate monolingual data in MTNT corpus. The back-translated data is combined with the noisy parallel data and used to fine-tune the baseline model. For Fr$\rightarrow$En, by introducing back-translated data, the model performance drops by over 2 BLEU scores compared the simply tuning on parallel data. This would suggest that the back translation data might break the noise level gap between source and target texts, and hence the model fine-tuned on back-translated data tends to output noisier translations and performs worse.

Since the size of the MTNT dataset is too small, we tried merging the data in both language directions, resulting in 55k sentence pairs for fine tuning. For both directions, the models tuned on the merged MTNT data (Tune-B) show worse performance than the models tuned on single direction data (Tune-S). This is due to the introduction of opposite direction data would increase the noise in target texts. We added the forward translation and fuzzy matches data separately and fine-tuned with the merged MTNT data. Results show that the introduction of either forward translation or fuzzy match data would improve model performance on noisy test sets, compared to the Tune-B model. However, with only forward translation or fuzzy match data added, the model still lags behind the Tune-S performance. Therefore, we mix the FT, FM, and merged MTNT data. After we use the mixed data for fine tuning, models in both directions scored better with the augmented data. The Fr$\rightarrow$En model with forward translation and fuzzy matches data achieved a performance of 42.80 BLEU score on the MTNT2019 test set, an improvement of +1.32 BLEU points over the Tune-S model. The forward translation data is generated using the Tune-B model, which includes information on the opposite direction, and this might benefit forward translation and prevent the model from overfitting. Compared to back translation, forward translation could keep the noise level difference between the source and target sentences\footnote{We experimented with fuzzy matches plus MTNT data (not merged), but it does not improve performance, because without the opposite direction information the model overfits to the tuning data.}. 

\subsection{Double Fine Tuning}
Considering that the opposite direction data from that of the MTNT dataset would harm the model performance, we applied  double fine tuning to compensate. We used the model which had already been tuned on the combination of forward translation data, fuzzy match data and merged MTNT data, and fine-tuned it with the MTNT data on the corresponding direction (e.g. Fr$\rightarrow$En data for Fr$\rightarrow$En model). In this case, the MTNT data in the same language direction would fine-tune the model twice, thus adapting the model to the specific language direction domain. The second fine tuning was able to further improve model robustness to noisy data and keep a similar performance on clean data. In the En$\rightarrow$Fr direction, the second fine tuning improves +0.57 and +0.85 BLEU points on MTNT and MTNT2019. 

\subsection{Punctuation Fixing}
The MTNT2019 test set uses a different set of punctuation in French text as the MTNT dataset and clean training data. In the MTNT2019 test set, the French references use apostrophes (') and angle quotes (« and »), instead of the single quotes (\textquotesingle) and double quotes (") used in the MTNT training data \cite{berard-calapodescu-roux:2019:WMT}. Therefore, models fine-tuned with MTNT training data would show an inconsistent performance for punctuation when evaluating on the MTNT2019 test set. We fixed the punctuation in the En$\rightarrow$Fr direction as a postprocessing step. This single replacement improves +4.36 BLEU score over the double fine-tuned model. For comparison, we also postprocessed the output from the Tune-S model. The punctuation fix results in an increase of +3.24 BLEU score.

\subsection{External Data}
To explore the effect of other types of noise, we fine-tuned our baseline model on different external datasets (see the ``Unconstrained" rows in Table \ref{table:bleu_fren} and \ref{table:bleu_enfr}). We experimented with human transcript and translation in IWSLT dataset. The BLEU score (Fr$\rightarrow$En) on MTNT2019 increases by +2.14 over the baseline, while the results on the other three test sets decrease. In the En$\rightarrow$Fr direction, fine tuning on IWSLT improves the model performance on all four test sets, and with MTNT data added, the BLEU score on noisy data performs even better than the Tune-B model. The benefit of speech transcripts might come from informal languages such as slang, spoken language, and domain-related words. Apart from this, we also kept the indicating words (e.g. ``[laughter]" and ``[applause]") in the transcripts, which could also play a role of noise. 

When using ASR data generated from the audio files in the MuST-C dataset, we first filtered ASR data by removing sentences where the original transcript length is over 1.5 times that of the ASR transcript. The model fine-tuned on ASR data shows a slight decrease in BLEU scores. We found that the ASR transcript often skips some phrases in a sentence. Therefore, we reduced the length ratio threshold to 1.2, and with that the model achieves similar performance as the baseline model. Evaluated on \textit{newstest}, the ASR-tuned model improves +1.08 BLEU score over the baseline. Finally, we tried with the parallel texts from human transcript and translation in MuST-C corpus, similar to IWSLT, the performance increases on all test data. This suggests that the introduction of external data with different types of noise could improve model robustness, even without the use of in-domain noisy data.

Finally, we conducted a domain-sensitive training experiment by adding tags for different data. We mixed all available data, including the MTNT data (two directions), IWSLT, MuST-C, ASR, forward translation, and fuzzy match data. Tags are added at the beginning of the source sentences. As shown in the last row in Table \ref{table:bleu_enfr},  mixed training could improve the performance over baseline on noisy texts. However, the model does not outperform the models with fine tuning. This might result from the introduction of ASR generated data, which can contain more low quality training samples.

\subsection{WMT19 Robustness Leaderboard}
We submitted our best constrained systems to WMT19 Robustness Leaderboard\footnote{\url{http://matrix.statmt.org}}, as shown in Table \ref{tab:board-fren} and Table \ref{tab:board-enfr}. In the Fr$\rightarrow$En direction, we submitted the model fine-tuned on merged MTNT data, forward translation and fuzzy match data (row 5 in Table \ref{table:bleu_fren}). In the En$\rightarrow$Fr direction, the double-tuned model with punctuation fixed was submitted (row 6 in Table \ref{table:bleu_enfr}).
\begin{table}[htb]
\centering
\begin{tabular}{|l|c|}
\hline
\textbf{System} & \textbf{BLEU-uncased}\\ \hline
\cite{berard-calapodescu-roux:2019:WMT} & 48.8 \\ \hline
\cite{helcl-libovick-popel:2019:WMT} & 45.8 \\ \hline
\cite{zheng-EtAl:2019:WMT} & 44.5 \\ \hline
Ours & 43.8 \\ \hline
\cite{post-duh:2019:WMT} & 41.8 \\ \hline
\cite{zhou-EtAl:2019:WMT} & 36.0 \\ \hline
\cite{grozea:2019:WMT} & 30.8 \\ \hline
MTNT paper baseline & 26.2 \\ \hline
\end{tabular}
\caption{WMT19 Robustness Leaderboard on Fr$\rightarrow$En.}
\label{tab:board-fren}
\end{table}

\begin{table}[htb]
\centering
\begin{tabular}{|l|c|}
\hline
\textbf{System} & \textbf{BLEU-uncased}\\ \hline
\cite{berard-calapodescu-roux:2019:WMT} & 42.0 \\ \hline
\cite{helcl-libovick-popel:2019:WMT} & 39.1 \\ \hline
Ours & 37.1 \\ \hline
\cite{zheng-EtAl:2019:WMT} & 37.0 \\ \hline
\cite{grozea:2019:WMT} & 24.8 \\ \hline
MTNT paper baseline & 22.5 \\ \hline
\end{tabular}
\caption{WMT19 Robustness Leaderboard on En$\rightarrow$Fr.}
\label{tab:board-enfr}
\end{table}
Our systems would have achieved the 4th and 3rd place on Fr$\rightarrow$En and En$\rightarrow$Fr directions. The leading systems use back translation on a large volume of clean monolingual data, therefore could benefit from the size of clean data. Although our system does not utilize clean monolingual data, we find an alternative way to extend noisy parallel data, which might be more efficient for training. The results show that our systems could achieve a competitive position.

\section{Conclusions}
In this paper we use data augmentation strategies to improve neural machine translation models robustness. We experiment under the setting of the WMT19 Robustness Task for the Fr$\leftrightarrow$En language directions. We propose the use of forward translation and fuzzy matches as alternatives to back translation to augment noisy data. Our best models with augmented noisy data could improve +1.32 and +0.97 BLEU scores for Fr$\rightarrow$En and En$\rightarrow$Fr  over models fine-tuned with noisy parallel data. We also explore the effect of external noisy data in the form of speech transcripts and show that models could benefit from data injected with noise through manual transcriptions of spoken language. The ASR generated data does not help improving robustness as it contains low quality training samples that break the sentences, while the human transcripts from speech proved helpful to translate noisy texts, even without in-domain data. Future work might include training a domain-related speech recognition model and generate better ASR parallel data instead of using the off-the-shelf system.



\bibliography{emnlp-ijcnlp-2019}
\bibliographystyle{acl_natbib}

\end{document}